\documentclass[11pt, letterpaper]{article}

\usepackage[margin=1in]{geometry}
\usepackage{amsmath, amssymb, amsfonts}
\usepackage{graphicx}
\usepackage{booktabs}
\usepackage{multirow}
\usepackage{caption}
\usepackage{subcaption}
\usepackage{adjustbox}
\usepackage{makecell} 
\usepackage{xcolor}
\usepackage{url}
\usepackage[colorlinks=true, linkcolor=blue, citecolor=blue, urlcolor=blue]{hyperref}

\usepackage[numbers,sort&compress]{natbib}
\newcommand{\wellfigdir}{.}
\newcommand{\welltraj}{traj00}
\newcommand{\wellmode}{single}
\newcommand{\wellpath}[2]{\wellfigdir/#1/#1__#2__\welltraj__\wellmode}
\newcommand{\wellfig}[3]{%
  \begin{figure}[htbp]
    \centering
    \includegraphics[width=\linewidth]{\wellpath{#1}{#2}.pdf}
    \caption{#3}
    \label{fig:#1:#2}
  \end{figure}%
}

\newcommand{\gtfno}[2]{#1, #2: ground truth (top) and tuned-TFNO prediction (bottom).}

\newcommand{\rollout}[2]{%
\begin{figure}[ht]
\centering
\IfFileExists{figures/#1.pdf}%
{\includegraphics[width=\textwidth]{figures/#1.pdf}}%
{\fbox{\parbox[c][3.0cm][c]{0.95\textwidth}{\centering\textcolor{red}{%
[figure placeholder: \texttt{figures/#1.pdf} --- ground truth, tuned-FNO prediction, and absolute error at a representative rollout step]}}}}
\caption{#2}
\label{fig:#1}
\end{figure}}

\title{Performance of Neural Operator Baselines on `The Well' Dataset}
\author{Yash Pincha, Andy Dimnaku, Andrew Chiang, Adarsh Ganeshram \\ Valentin Duruisseaux, Anima Anandkumar 
}
\date{}

\begin{document}

\maketitle

\begin{abstract}
\noindent \emph{The Well}, a 15-terabyte dataset spanning 16 physical simulation domains, provides an extensive benchmark for developing and evaluating machine learning surrogate models for complex physical systems. The baselines released with the benchmark were presented by their authors as illustrative examples obtained at small computational cost, trained with zero to very limited hyperparameter tuning and under a highly constraining 12-hour per-run wall-clock limit. These results have subsequently been treated as fully tuned benchmarks, and in some cases effectively interpreted as unconstrained performance baselines. We perform a short non-exhaustive hyperparameter search for the Tucker-factorized Fourier Neural Operator (TFNO) under the same 12-hour per-run wall-clock limit, while using only a single input timestep rather than the history of up to four used by the published baselines. Despite this more restrictive setup, tuning alone reduces VRMSE below both published FNO and TFNO baselines on 15 of 17 tasks, sometimes by several orders of magnitude. This sensitivity to a modest tuning effort underscores that the released neural operator baselines should not be treated as architectural benchmarks. 
\end{abstract}
\section{Introduction}

The simulation of high-dimensional, spatiotemporal physical systems using traditional numerical solvers imposes prohibitive computational costs, often requiring millions of core hours for high-resolution fluid dynamics, magnetohydrodynamics, and astrophysical phenomena. Machine learning surrogate models offer a compelling alternative, predicting the evolution of physical systems at a fraction of the computational cost of traditional numerical solvers. Among these approaches, neural operators learn mappings between infinite-dimensional function spaces~\cite{kovachki2023neural,azizzadenesheli2024neural,Berner2025Principles}, making them particularly well suited to approximating solution operators for complex physical systems.

\paragraph{The  Well.} \citet{TheWell} recently introduced \emph{The Well}, a large-scale collection of diverse physics simulations designed to standardize the evaluation of surrogate models. Curating a 15-terabyte benchmark spanning 16 scientific domains is a significant contribution to both the machine learning and computational physics communities, and complements earlier efforts such as PDEBench~\cite{takamoto2022pdebench}. 

\paragraph{The concern.} The baselines reported by \citet{TheWell} were explicitly presented as performance examples obtained at small computational cost and not optimized results, as they were therefore trained under a largely shared configuration with zero to very limited model-specific tuning and a strict 12-hour per-run wall-clock limit. Our concern is that the resulting values have nevertheless been reused as baselines for comparison in subsequent work. If modest, dataset-specific tuning can substantially improve these values, then they do not adequately characterize the performance of the architecture, motivating our effort to quantify this sensitivity and assess whether the published baselines are sufficiently stable for the comparisons in which they have been reused.

\paragraph{In this paper.} We study Fourier neural operators, focusing on FNO/TFNO architectures~\citep{li2020fno,Kossaifi2023MGTFNO}, using the \texttt{neuraloperator} implementation~\cite{kossaifi2024neural}. Our experiments use the Tucker-factorized TFNO parameterization and a separate small hyperparameter search for each dataset under the 12-hour training limit. We make no other architectural modifications and use no data augmentation, pretraining, ensembling, or rollout-specific objectives. Our evaluation protocol is also more restrictive than that of the published baselines. \emph{The Well} conditions on up to four previous timesteps, whereas our models use a single input timestep to predict a single output timestep. Despite this reduced temporal context, modest hyperparameter tuning is sufficient to achieve substantially better results.\\

\noindent Under this more restrictive single-timestep protocol, the lightly tuned TFNO improves on both published neural operator baselines on 15 of 17 tasks. The improvement is highly dataset-dependent, more than an order of magnitude on several tasks, and exceeding $100\times$ on one dataset. Many runs are still improving when the 12-hour budget expires, and increasing the budget to 36 hours on an example dataset significantly reduces VRMSE by a further factor of $4.7$. These results establish that the published tables are not stable under even a modest amount of dataset-specific tuning, and therefore should not be used as architectural benchmarks. The code accompanying this study is publicly available at \href{https://github.com/neuraloperator/the_well_baselines}{github.com/neuraloperator/the\_well\_baselines}.

\vspace{6mm}

\subsection{Scope and Interpretation of the Results}
\label{sec:claims}

\vspace{4mm}

\begin{enumerate}
\item \textbf{We do not claim optimality.} Our search covers a few configurations per dataset, uses a single random seed, and explores only a coarse subset of hyperparameter space. We do not vary optimizer family, weight decay, normalization, loss function, or activation. The selected configurations are thus not claimed to represent fully optimized TFNO performance.

\vspace{2mm}

\item \textbf{We do not compare model families.} Only TFNO is tuned in our experiments. The convolutional baselines in \emph{The Well} were evaluated under a largely shared protocol with little to no tuning and may therefore also be far from their best performance. Comparisons with our tuned TFNO should not be interpreted as evidence of architectural superiority and are reported mainly for continuity with the original benchmark. We nevertheless expect the tuning gains to transfer to FNO. In \texttt{neuraloperator}, TFNO shares the FNO architecture but uses a Tucker-factorized parameterization for the spectral weights. At high rank, it is close to a dense FNO in parameter count and flexibility. Since most tuned hyperparameters, including the number of modes, width, depth, batch size, and learning rate, are shared by both models, comparable tuning would also improve the FNO baseline.

\vspace{2mm}

\item \textbf{The 12-hour training regime.} The wall-clock limit constrains both training and model selection. A configuration that is larger, slower per step, or paired with a gentler learning-rate schedule receives fewer useful updates before the clock expires and is therefore less likely to be selected, even if it would attain lower error with more time. Conversely, small models and aggressive schedules are rewarded for reaching acceptable validation error quickly. The configurations reported here are thus outcomes of a coarse search conducted inside the original 12-hour constraint, not principled estimates of either unconstrained TFNO performance or an optimum specifically tailored to 12 hours (see Section~\ref{sec:extended}).

\vspace{2mm}

\item \textbf{The claim.} Our claim concerns the reliability and sensitivity of the published neural operator numbers when reused as benchmarks. If the same FNO/TFNO implementation improves by up to two orders of magnitude after a small configuration search under the same nominal per-run wall-clock limit, then the published values are better understood as outcomes of a particular configuration and search protocol than as stable measures of architectural capability. Beating those values alone is therefore not evidence of architectural superiority. The quantitative content of this paper is the size of that gap.
\end{enumerate}

\vspace{6mm}

\section{Background and Methodology}

\subsection{\emph{The Well} Datasets}
\emph{The Well} provides approximately 15~TB of simulation data across 16 distinct scientific domains. The phenomena captured range from biological active matter to complex, multi-scale astrophysical flows. Individual datasets range in size from 6.9~GB to 5.1~TB. All data is standardized on uniform spatial grids, sampled at constant time intervals, and stored in HDF5 format to facilitate scalable machine learning workflows. We evaluate on 17 predictive tasks derived from the 16 base datasets, using the canonical train/validation/test splits; per-dataset descriptions and qualitative rollouts are deferred to Appendix~\ref{app:datasets}.

\subsection{Neural Operators and the FNO}
\label{sec:fno}

Neural operators learn mappings between infinite-dimensional function spaces, so that a single trained model can in principle be evaluated at resolutions and on discretizations other than those it was trained on~\cite{kovachki2023neural,Berner2025Principles}. Given an input function $a$ defined on a domain $D$, a neural operator $\mathcal{G}_\theta$ lifts $a$ pointwise to a higher-dimensional latent representation $v_0$, applies a sequence of operator layers, and projects the result back to the output function space. Each layer combines a pointwise linear map with a non-local integral operator,
\begin{equation}
v_{l+1}(x) = \sigma\!\left( W v_l(x) + b + (\mathcal{K}_l v_l)(x) \right), \qquad x \in D,
\end{equation}
where $\sigma$ is a nonlinearity and $\mathcal{K}_l$ is a kernel integral operator. The non-local term is what distinguishes the construction from a pointwise network, and different choices of $\mathcal{K}_l$ recover different architectures. The Fourier Neural Operator (FNO)~\cite{li2020fno} parametrizes $\mathcal{K}_l$ as a spectral convolution,
\begin{equation}
(\mathcal{K}_l v_l)(x) = \mathcal{F}^{-1}\!\left( R_l \cdot \mathcal{F}(v_l) \right)(x),
\end{equation}
where $\mathcal{F}$ is the (discrete) Fourier transform and $R_l$ is a learned complex weight tensor applied to the lowest $k_{\max}$ Fourier modes, with higher modes truncated. The retained mode count controls the spectral bandwidth on which the learned convolution can act and is therefore a central FNO hyperparameter. Too few modes can suppress physically relevant fine-scale structure, whereas retaining too many relative to the grid resolution increases cost and can expose subsequent nonlinear layers to unresolved high-frequency content and aliasing~\cite{FNOGuide}. Because each retained frequency carries learned channel-mixing weights, increasing the mode count also enlarges the spectral parameter tensor. The Tensorized FNO (TFNO)~\cite{Kossaifi2023MGTFNO} reduces this cost by factorizing that tensor, most commonly with a Tucker decomposition~\cite{kossaifi2019tensorly}, whose rank ratio controls the degree of compression, ranging from strongly factorized representations to the dense limit.

\noindent To keep the software implementation fixed, all models in this study use the FNO and TFNO implementations in the \texttt{neuraloperator} library~\cite{kossaifi2024neural}, the same library used for the neural operator baselines in \emph{The Well}. The comparison therefore changes model and training configuration without introducing a separate FNO codebase. The models are trained with the Adam optimizer~\cite{kingma2015adam} under a cyclic learning-rate schedule~\cite{smith2017cyclical}. Each run is limited to 12 hours on a single H100 GPU and uses a fixed random seed of 42. The dataset-specific hyperparameters varied in our search, including the number of layers, Fourier modes, channel width, batch size, learning rate, Tucker rank, and scheduler patience, are reported in the appendix in Table~\ref{tab:hyperparameters}.

\subsection{Temporal Conditioning Protocol}
\label{sec:protocol}

A methodological difference between this study and the original benchmark concerns the extent of the trajectory the model is permitted to observe. The baselines reported in \emph{The Well} are trained autoregressively with a temporal history window: the model receives up to four consecutive prior states, $(u_{t-3}, u_{t-2}, u_{t-1}, u_t)$, stacked along the channel dimension, and is trained to predict $u_{t+1}$. Access to multiple prior frames provides additional information about the system's temporal evolution, including information from which local finite-difference estimates can be formed. With four frames, higher-order temporal structure is also available, which materially simplifies the learning problem for advection-dominated and oscillatory systems.\\

\noindent Each TFNO model reported here receives only a single state $u_t$ and predicts $u_{t+1}$ directly. No temporal history is provided during training or inference, so the operator must infer from the instantaneous spatial field alone. We adopt this more restrictive protocol to ensure that any improvement cannot be attributed to additional input information and to standardize future searches. Consequently, the comparison in Section~\ref{sec:results} is deliberately unfavorable to the TFNO models trained here, and a tuned TFNO given the full four-step history would be expected to perform better.

\subsection{Evaluation Metric}
To quantify performance consistently across domains with substantially different physical scales and state variables, the benchmark employs the Variance-scaled Root Mean Square Error (VRMSE). VRMSE normalizes the root mean square error by the standard deviation of the ground-truth field, which guarantees that a trivial surrogate predicting only the spatiotemporal mean of the target variable attains a score of exactly $1.0$. The metric is defined as
\begin{equation}
\text{VRMSE}(y, \hat{y}) = \sqrt{ \frac{\mathbb{E}\left[(y - \hat{y})^2\right]}{\text{Var}(y)} }
\end{equation}
where $y$ denotes the true state of the physical system, $\hat{y}$ denotes the surrogate model's prediction, and the expectation is taken over the entire spatiotemporal grid. Lower values indicate greater predictive accuracy, with a theoretical minimum of zero.

\subsection{Extent of the Hyperparameter Search}
\label{sec:search}

The scale of the search is important for interpreting the results. We vary seven hyperparameters over a coarse, manually guided grid, evaluating on the order of tens of configurations per dataset and selecting models by VRMSE. The candidate values are listed in Appendix~\ref{app:grid}.\\

\noindent All other settings, listed in the caption of Table~\ref{tab:hyperparameters}, are fixed across the 17 tasks, including choices that a more systematic study would typically tune, such as optimizer family, learning-rate schedule, lifting and projection widths, and domain padding. Each selected configuration is trained only once, so no variance estimates are available and some of the smaller differences may reflect seed variability. The U-net and CNextU-net baselines were taken from The Well's numbers. This deliberately limited search is not intended to establish optimal TFNO configurations.\\

\noindent By the standards of the hyperparameter optimization literature, this is a relatively limited search. Nevertheless, the substantial improvements over the published baseline figures, in some cases by one or two orders of magnitude, demonstrate that these values do not fully characterize the performance attainable by the architecture. A more systematic search could yield further improvements.

\vspace{8mm}

\section{Results and Discussion}
\label{sec:results}

\vspace{1mm}

\subsection{Updated Performance}

\vspace{4mm}

\paragraph{Overall performance.}
Table~\ref{tab:optimized_results} reports VRMSE for the tuned TFNO alongside the four baselines published in \emph{The Well}. Under the same 12-hour per-dataset compute budget and using only a single input timestep, dataset-specific hyperparameter tuning achieves lower error than both published neural operator baselines on 15 of the 17 tasks, sometimes by one to two orders of magnitude.

\paragraph{Comparison with convolutional baselines.}
The U-net and CNextU-net results are reproduced only for continuity with the original benchmark. A comparable tuning study for the convolutional baselines remains to be carried out. As \citet{TheWell} note for all of their baselines, these received little model-specific tuning, so comparisons between them and our tuned TFNO do not support conclusions about relative architectural performance.

\paragraph{Tasks without improvement.}
The two exceptions are the \texttt{convective\_envelope\_rsg} and \texttt{turbulence\_gravity\_cooling} datasets. Both 3D tasks trained with relatively small batches. For \texttt{turbulence\_gravity\_cooling}, the selected run reaches the 12-hour limit after only 13 epochs while validation error is still decreasing (Table~\ref{tab:hyperparameters}); its VRMSE of $0.2758$ is close to the published TFNO value of $0.2673$, but the run is plainly budget-truncated. \texttt{convective\_envelope\_rsg} triggers early stopping after 40 epochs and reaches $0.0309$, compared with the best published operator value of $0.0269$. As Section~\ref{sec:convergence} explains, however, early stopping under a schedule and patience chosen inside a fixed wall-clock budget should not be interpreted as budget-independent convergence. The small residual gaps on these two tasks are therefore not diagnostic of architectural limits: they remain entangled with the coarse search, the single-timestep conditioning protocol, stochastic variation, and the budget-shaped optimization schedule, none of which we attempt to separate.
\paragraph{Variation across tasks reveals protocol sensitivity.}
Table~\ref{tab:optimized_results} expresses the same comparison as ratios, making the dataset-to-dataset variation explicit. Relative to the reported FNO or TFNO values, the ratio ranges from $0.9\times$ to $>75\times$; for \texttt{rayleigh\_taylor\_instability}, where all four published baselines have VRMSE $>10$, the tuned TFNO value of $0.1333$ implies an improvement of at least $75\times$. If the published operator values were robust to reasonable configuration choices, a small search using the same implementation and the same per-run compute limit would be expected to leave most ratios near unity. Instead, the difference varies by more than two orders of magnitude across tasks. That spread is direct evidence that the published point estimates are strongly protocol-dependent.

\paragraph{Sensitivity to hyperparameter choice.}
The response to dataset-specific tuning is highly heterogeneous. The largest operator-baseline displacements occur on \texttt{rayleigh\_taylor\_instability} ($>75\times$), \texttt{planetswe} ($27\times$), and \texttt{acoustic\_scattering\_maze} ($22\times$), whereas several 3D astrophysical tasks remain close to the published values. Because hyperparameters are varied jointly, these differences cannot be assigned to any one change. Several of the modest-gain 3D runs are also among the most compute-constrained, further preventing a clean causal interpretation. The relevant conclusion is therefore not that one particular setting fixes the baselines, but that the distance between a shared configuration and a lightly tuned configuration depends strongly on the dataset. A common setting can perform reasonably well on one task while yielding substantially less favorable results on another, so a single shared protocol does not reliably characterize TFNO performance across this heterogeneous collection.

\vspace{8mm}

\begin{table}[h!]
\centering
\small
\setlength{\tabcolsep}{3pt}
\caption{Sensitivity of the reported operator baselines to protocol and configuration. Column~2 gives the VRMSE of our 12h-tuned TFNO. Columns~3 and~4 give the ratio of the published TFNO and FNO VRMSEs from \emph{The Well}~\cite{TheWell} to ours, each computed as \emph{published value divided by tuned value}; column~5 gives the same ratio against the best (lowest) of all four published baselines on that dataset. Values above one indicate that our run scores lower than the reference; values below one indicate that it scores higher. The two baselines are reported separately rather than aggregated, since on \texttt{gray\_scott\_reaction\_diffusion} and \texttt{helmholtz\_staircase} the published FNO substantially outperforms the published TFNO, and any single aggregate would obscure this.}
\label{tab:optimized_results}
\begin{tabular}{lcccc}
\toprule
\textbf{Dataset} &
\textbf{12h-tuned} &
\textbf{vs.\ pub.\ TFNO} &
\textbf{vs.\ pub.\ FNO} &
\textbf{vs.\ Best Reported} \\
\midrule
\texttt{acoustic\_scattering\_maze}        & 0.022886 & $22.1\times$  & $22.1\times$  & $0.7\times$ \\
\texttt{active\_matter}                     & 0.077859 & $4.6\times$   & $4.7\times$   & $1.3\times$ \\
\texttt{convective\_envelope\_rsg}          & 0.030917 & $0.9\times$   & $0.9\times$   & $0.9\times$ \\
\texttt{euler\_multi\_quadrants}            & 0.060337 & $6.9\times$   & $6.8\times$   & $2.5\times$ \\
\texttt{gray\_scott\_reaction\_diffusion}   & 0.022986 & $15.8\times$ & $5.9\times$  & $5.9\times$ \\
\texttt{helmholtz\_staircase}               & 0.000280 & $12.5\times$  & $1.8\times$   & $1.8\times$ \\
\texttt{MHD\_64}                            & 0.212129 & $1.7\times$   & $1.7\times$   & $0.8\times$ \\
\texttt{planetswe}                          & 0.003143 & $27.1\times$  & $54.9\times$  & $27.1\times$ \\
\texttt{post\_neutron\_star\_merger}        & 0.369881 & $1.0\times$   & $1.0\times$   & $1.0\times$ \\
\texttt{rayleigh\_benard}                   & 0.154014 & $4.3\times$   & $5.5\times$   & $4.3\times$ \\
\texttt{rayleigh\_taylor\_instability}      & 0.133326 & $>75.0\times$ & $>75.0\times$ & $>75.0\times$ \\
\texttt{shear\_flow}                        &  0.085620 & $17.3\times$   & $14.1\times$   & $9.6\times$ \\
\texttt{supernova\_explosion\_64}           & 0.342370 & $1.1\times$   & $1.1\times$   & $0.9\times$ \\
\texttt{turbulence\_gravity\_cooling}       & 0.275790 & $1.0\times$   & $0.9\times$   & $0.8\times$ \\
\texttt{turbulent\_radiative\_layer\_2D}    & 0.220964 & $2.2\times$   & $2.2\times$   & $0.9\times$ \\
\texttt{turbulent\_radiative\_layer\_3D}    & 0.309779 & $1.7\times$   & $1.7\times$   & $1.2\times$ \\
\texttt{viscoelastic\_instability}          & 0.095204 & $7.5\times$   & $7.6\times$   & $2.6\times$ \\
\end{tabular} \vspace{4mm}
\end{table}

\paragraph{Dataset-specific configurations.}
Table~\ref{tab:hyperparameters} reports the selected configuration for each dataset. The substantial variation across tasks is itself informative: Tucker rank ranges from $0.05$ on \texttt{convective\_envelope\_rsg} to $1.0$ on five tasks, spectral mode counts range from 32 to 256, and batch sizes from 1 to 64. No single assignment of these parameters is adequate across all datasets. This variation illustrates why a largely shared configuration cannot reliably characterize TFNO performance across heterogeneous datasets. The selected values are not claimed to be optimal, and several were chosen from only three or four candidate values.\\

\subsection{Termination Status and Compute Saturation}
\label{sec:convergence}

Table~\ref{tab:hyperparameters} reports both the number of completed epochs and how each selected run terminated. Seven of the seventeen runs exhaust the 12-hour wall-clock budget while validation error is still decreasing; those values are unambiguously unfinished estimates for the selected configurations. The other ten runs trigger the chosen early-stopping rule. This should not be read as evidence of budget-independent convergence. Both the learning-rate schedule and the stopping patience were selected inside the same 12-hour constraint, so an early-stopped run has only reached the endpoint induced by that compressed training protocol. A longer budget can support a different schedule, a different patience, or a different configuration altogether, as Section~\ref{sec:extended} demonstrates on one task. The ratios in Table~\ref{tab:optimized_results} carry the same qualification and should not be interpreted as fully optimized TFNO performance.

\paragraph{Budget-halted runs are lower bounds.}
For the seven budget-halted runs the reported values are, at least, clear lower bounds on the accuracy attainable by the stated configuration, and the corresponding entries in Table~\ref{tab:optimized_results} are lower bounds on the sensitivity being measured. Several of the largest gains arise among these runs: \texttt{gray\_scott\_reaction\_diffusion} reaches $15.8\times$ after 7 epochs, \texttt{acoustic\_scattering\_maze} $22\times$ after 17, \texttt{euler\_multi\_quadrants} $6.9\times$ after 4, and \texttt{rayleigh\_benard} $4.3\times$ after 9. That improvements of this size appear within the first several epochs indicates they come from the configuration itself rather than from extended optimization, and these gains would be expected to increase significantly with additional training.

\paragraph{Equal compute does not imply equal optimization.}
The completed epoch count ranges from 4 to 596, so equal wall-clock time does not provide equal amounts of optimization across datasets. The resulting VRMSE therefore mixes at least three effects: the quality of the configuration, its per-step computational cost, and how rapidly the chosen optimizer and schedule reduce error. The same issue applies to the published baselines, which were subject to the same wall-clock cap; because their termination states and training trajectories are not reported, one cannot tell which values were still improving when the budget ended.

\paragraph{Search limitation and budget bias compound.} Two distinct constraints keep the reported values from representing optimized TFNO performance. First, the configuration search is deliberately coarse and leaves many consequential choices unexplored. Second, every candidate is evaluated inside a wall-clock limit that favors configurations and schedules that improve quickly. A potentially strong configuration can therefore be missed either because it was never sampled or because its training dynamics make it unattractive under the 12-hour clock. The dedicated hyperparameter-tuning discussion in \citet{FNOGuide} reinforces that these choices are problem-dependent rather than universal defaults. Our results should therefore be read as evidence that a shared configuration is insufficient.
 
\vspace{6mm}

\subsection{Beyond the Fixed Budget}
\label{sec:extended}

To test whether the 12-hour cap is constraining, we raised it to 36 hours on one task and repeated the search of Section~\ref{sec:search} under the larger budget. We chose \texttt{acoustic\_scattering\_maze}, which under the original protocol was halted after 17 epochs with validation error still decreasing.

\vspace{4mm}

\begin{table}[ht]
\centering
\caption{Effect of increasing the per-run wall-clock budget from 12 to 36 hours on the dataset \texttt{acoustic\_scattering\_maze}. For each budget, the configuration and cyclic learning-rate schedule are re-selected, so the comparison captures the combined effect of additional training time and the different configurations that become competitive under the larger budget; it is not a continuation of the 12-hour run. Both runs are halted by their respective wall clocks while validation error is still decreasing and should therefore be treated as budget-limited. \textbf{vs.\ 12h} is $0.022886$ divided by the row's VRMSE. \textbf{vs.\ operator baseline} uses the larger of the published FNO and TFNO VRMSEs ($0.5062$) divided by the row's VRMSE. Ratios against the published operator values quantify how far those protocol-specific figures can be displaced and are not evidence of architectural superiority.}
\label{tab:extended}
\resizebox{\textwidth}{!}{
\begin{tabular}{lccccccc|cc}
\toprule
& \multicolumn{7}{c|}{\textbf{Configuration and outcome}} & \multicolumn{2}{c}{\textbf{Ratios}} \\
\cmidrule(lr){2-8} \cmidrule(lr){9-10}
\textbf{Budget} & \textbf{Modes} & \textbf{Width} & \textbf{Batch} & \textbf{LR} &
\textbf{Rank} & \textbf{Epochs} & \textbf{VRMSE} &
\textbf{vs.\ 12h} & \textbf{vs.\ operator baseline} \\
\midrule
12\,h & 64  & 32 & 32 & $2\times10^{-3}$ & 0.20 & 17 & 0.022886 & $1.0\times$ & $22.1\times$ \\
36\,h & 128 & 48 & 8  & $8\times10^{-4}$ & 1.00 & 38 & 0.004875 & $4.7\times$ & $103.8\times$ \\
\bottomrule
\end{tabular}
}
\end{table}

\vspace{4mm}

\paragraph{The budget changes the selected configuration.} The 36-hour search chooses a substantially more expensive configuration and a gentler optimization schedule. The retained mode count doubles from 64 to 128, the channel width increases from 32 to 48, and the Tucker rank rises from $0.20$ to $1.00$, the dense limit. At the same time, batch size decreases from 32 to 8, the learning rate from $2\times10^{-3}$ to $8\times10^{-4}$, and early-stopping patience increases from 8 to 15, with depth unchanged. The comparison therefore does not isolate ``more epochs'' for a fixed model. It shows that a larger wall-clock budget changes which region of the configuration space is competitive: higher-capacity models and slower schedules can receive enough optimization to be selected. This selection effect is precisely why a fixed wall-clock cap cannot be treated as neutral with respect to model configuration.

\paragraph{The reachable error continues to fall.} Increasing the budget reduces VRMSE from $0.022886$ to $0.004875$, a further factor of $4.7$, while the selected run completes 38 rather than 17 epochs. Relative to the published operator values ($0.5057$ for TFNO and $0.5062$ for FNO), the displacement grows from roughly $22\times$ at 12 hours to roughly $104\times$ at 36 hours. This is evidence of a binding constraint.

\paragraph{What the 12-hour protocol measures.} A fixed 12-hour cap measures performance under a joint constraint on model size, optimization speed, and training duration. It favors configurations that are inexpensive per step or improve rapidly, while configurations that require more computation per update or slower schedules are comparatively disadvantaged. It thus does not provide a general measure of what the architecture can attain on the dataset.\\

\subsection{Limitations}
\label{sec:limitations}

The experiments are intentionally sufficient to test the stability of the published neural operator baselines under modest tuning, but they are not sufficient to establish optimized TFNO performance. Four limitations are especially important.

\paragraph{Single-seed estimates.}
Each selected configuration is trained once with a fixed random seed. We therefore do not estimate run-to-run variability, and differences near unity in Table~\ref{tab:optimized_results} may be within ordinary stochastic variation. The largest displacements are much larger than the small differences for which seed variance is usually most consequential, but we do not empirically verify that assumption here. 

\paragraph{One-step error is not rollout quality.}
Following \emph{The Well}, the quantitative comparison uses single-step VRMSE. This preserves comparability with the published values but does not characterize long-horizon autoregressive stability, error accumulation, or physical consistency over a rollout. A configuration that minimizes one-step VRMSE need not be the configuration that produces the best long-horizon trajectory. A natural extension would be to test whether physics-informed neural operators~\cite{li2024pino,ganeshram2025fcpino,lin2025mGNO} can have improved long-horizon stability or physical consistency on \emph{The Well}. Physics-based losses can also reduce reliance on expensive high-resolution data by combining supervision on coarse grids with physics constraints evaluated at finer resolutions, while providing useful diagnostics of physical violations during rollout. Such comparisons should account for the additional supervision and computational cost of physics-informed training.

\paragraph{Published baselines are not independently reproduced.}
We compare against the values reported by \emph{The Well} rather than retraining every baseline. Our analysis therefore inherits the published training and evaluation protocol and cannot independently assess reproducibility, seed variance, termination state, or implementation details not captured in the reported results. This limitation is particularly relevant for small differences between methods.

\paragraph{The selected configurations are budget-conditioned.}
The reported TFNO configurations emerge from a coarse search conducted entirely under the 12-hour wall-clock constraint. They should not be interpreted as recommended TFNO settings, as estimates of unconstrained TFNO performance, or as replacement benchmark values. Section~\ref{sec:extended} shows on one task that relaxing the budget changes both the selected configuration and the attainable error; the extent of that effect has not been measured systematically across the other tasks.\\

\subsection{Implications for Benchmark Design on \emph{The Well}}

The results do not imply that \emph{The Well} is unsuitable as a dataset for comparative evaluation. They imply that the current baseline point estimates are insufficient as architectural benchmarks. 

\paragraph{Treat tuning as part of the experimental method.}
A final VRMSE cannot be interpreted independently of the search that produced it. Reports should specify which hyperparameters were varied, the candidate ranges or search procedure, the selection metric, the number of configurations evaluated, and the compute spent on search as well as on the selected run.

\paragraph{Equalize tuning effort across compared models.}
Architectural claims require comparable opportunities for each model family to reach a strong configuration. A carefully tuned new method should not be compared against a baseline evaluated under a largely shared setting with zero to very limited model-specific tuning. Table~\ref{tab:optimized_results} shows that this asymmetry can change the apparent gap by orders of magnitude. This applies equally to the values reported here.

\paragraph{Report optimization state along with final error.}
When a wall-clock cap is used, every result should include the number of epochs or optimization steps, the termination condition, and preferably the validation trajectory. A model halted while error is still falling is qualitatively different from one whose stopping criterion has fired, even though neither should automatically be interpreted as globally or budget-independently converged.

\paragraph{Name the compute-constrained quantity explicitly.}
If the scientific question is performance under a fixed compute budget, that budget should be part of the metric description and the resulting claim. ``VRMSE after at most 12 hours under search protocol $P$'' is a reproducible quantity; ``FNO performance on \emph{The Well}'' is a much broader claim that the present protocol does not support. Where feasible, budget-constrained results should be complemented by longer-budget runs to show whether rankings or configurations are stable as the constraint is relaxed.

\paragraph{Quantify stochastic variability.}
Headline comparisons should use multiple random seeds and report an uncertainty estimate or distribution, especially when model differences are small. The present study does not meet this standard and therefore should not be used to resolve close rankings.

\paragraph{Use qualified bounds when training is incomplete.}
For a run halted by the budget while validation error is still decreasing, the observed VRMSE should be reported as a budget-limited point rather than as the attainable error of the configuration. Correspondingly, an improvement ratio against a fixed baseline is only a lower bound on the ratio that might be reached with further successful optimization. This qualification is more informative than presenting an unfinished run as a definitive score.

\paragraph{Interpreting protocol-specific results.}
Neither the original neural operator values nor the tuned values reported in this paper should serve as benchmark targets for claims of architectural improvement. They are outcomes of stated configurations, search procedures, temporal-conditioning choices, seeds, and compute budgets. A meaningful benchmark on \emph{The Well} must compare models under sufficiently balanced and transparent protocols; until then, beating either table establishes only that one protocol produced a lower number than another, not that one architecture is superior.\\
    
\section{Conclusion}

Under the original 12-hour per-run wall-clock limit, and with only one input timestep rather than the history of up to four available to the published baselines, a small per-dataset TFNO search improves on both reported neural operator baselines on 15 of 17 tasks. The size of the change is highly non-uniform, ranging from essentially no improvement on a few tasks to more than two orders of magnitude on others. Seven selected runs are still improving when the 12-hour clock expires, and completed epoch counts range from 4 to 596. On \texttt{acoustic\_scattering\_maze}, increasing the budget to 36 hours changes the selected configuration and reduces VRMSE by another factor of $4.7$, while the longer run is itself still budget-limited.\\

\noindent These observations show that the published neural operator point estimates should not be interpreted as benchmarks of architectural capability, consistent with how \citet{TheWell} originally presented them. When the same implementation can improve by one or two orders of magnitude under modest dataset-specific tuning, the reported values cannot be interpreted as reliable measures of architectural capability. Consequently, outperforming the published \emph{The Well} neural operator baselines is not, by itself, sufficient evidence of architectural progress. Such comparisons must also account for differences in tuning effort, computational budget, temporal context, and stochastic replication.\\

\noindent The numbers reported in this paper should not replace the original values as a new benchmark table. Our search is coarse, single-seeded, and deliberately budget-constrained; several runs are unfinished, and the extended-budget experiment shows that at least one reported value continues to move substantially as the constraint is relaxed. Neither the original neural operator baselines nor our tuned TFNO results should therefore be used as definitive reference targets for benchmarking. \emph{The Well} remains a valuable dataset and evaluation suite, but credible architectural benchmarking on it requires a different methodology: comparable tuning budgets across model families, multiple seeds, explicit search and compute accounting, transparent termination status and training trajectories, and clear separation between performance under a specified resource-constrained protocol and claims about an architecture itself. Until those conditions are met, results on these tasks should be reported as protocol-conditioned measurements rather than as model benchmarks. 

\hfill \\

\clearpage

\appendix

\section{Hyperparameter Search Grid}
\label{app:grid}

For completeness, this appendix reports the search space, selected configuration, and termination outcome for each task. The candidate values for each varied hyperparameter are listed below. The grid was explored manually rather than exhaustively, with choices for each dataset informed by results observed on others.

\begin{table}[ht]
\centering
\caption{Candidate values per varied hyperparameter. All other settings were fixed to the values in the caption of Table~\ref{tab:hyperparameters}. The number of spectral modes was chosen separately for each dataset based on resolution.}
\label{tab:grid}
\begin{tabular}{ll}
\toprule
\textbf{Hyperparameter} & \textbf{Candidate values} \\
\midrule
Layers          & 4, 6 \\
Spectral modes  & 32, 64, 128, 256 \\
Channel width   & 24, 30, 32, 40, 48, 50, 64, 96 \\
Batch size      & 1--64, set by memory limits per dataset \\
Learning rate   & $1\times10^{-4}$, $5\times10^{-4}$, $1\times10^{-3}$, $2\times10^{-3}$ \\
Tucker rank     & 0.05, 0.10, 0.20, 0.50, 1.00 \\
Patience        & 3, 4, 6, 8 \\
\bottomrule
\end{tabular}
\end{table}

\hfill

\noindent The selected configurations vary substantially across tasks and are reported here for reproducibility. As emphasized in Section~\ref{sec:claims}, they are the outcomes of a limited, budget-constrained search rather than recommended or optimal TFNO settings. Note that at high rank, the TFNO becomes close to a dense FNO in parameter count and flexibility. In particular, rank $1.0$ has approximately the same number of parameters, although it remains factorized.

\begin{table}[ht]
\centering
\caption{Selected TFNO configurations and training outcomes for the 17 reported 12-hour runs. All runs use \texttt{model=FNO}, \texttt{factorization=Tucker}, \texttt{optimizer=Adam}, \texttt{lr\_method=cycle}, \texttt{normalize=1}, \texttt{lifting\_channels=projection\_channels=64}, and \texttt{max\_train\_hours=12}. \textbf{Epochs} is the number completed before termination. \textbf{ES} marks whether the early-stopping criterion fired within the budget (\checkmark) or the run was halted by the 12-hour wall clock while validation error was still decreasing ($\times$). A \checkmark\ records only the behavior of the budget-shaped schedule and is not evidence of budget-independent convergence (Section~\ref{sec:convergence}). A rank of 1.00 denotes the dense limit. Configurations were selected by the limited search described in Section~\ref{sec:search} and are not claimed to be optimal.}
\label{tab:hyperparameters}
\resizebox{\textwidth}{!}{
\begin{tabular}{l c c c c c c c c c}
\toprule
\textbf{Dataset} &
\textbf{Layers} &
\textbf{Modes} &
\textbf{Width} &
\textbf{Batch Size} &
\textbf{LR} &
\textbf{Rank} &
\textbf{Patience} &
\textbf{Epochs} &
\textbf{ES} \\
\midrule
\texttt{acoustic\_scattering\_maze}        & 4 & 64  & 32 & 32 & $2\times10^{-3}$ & 0.20 & 8 & 17  & $\times$ \\
\texttt{active\_matter}                     & 6 & 128 & 48 & 32 & $1\times10^{-3}$ & 0.20 & 4 & 171 & \checkmark \\
\texttt{convective\_envelope\_rsg}          & 6 & 64  & 24 & 1  & $5\times10^{-4}$ & 0.05 & 6 & 40  & \checkmark \\
\texttt{euler\_multi\_quadrants}            & 6 & 128 & 30 & 12 & $5\times10^{-4}$ & 1.00 & 4 & 4   & $\times$ \\
\texttt{gray\_scott\_reaction\_diffusion}   & 6 & 64  & 64 & 32 & $5\times10^{-4}$ & 0.10 & 8 & 7   & $\times$ \\
\texttt{helmholtz\_staircase}               & 6 & 128 & 64 & 8  & $5\times10^{-4}$ & 0.10 & 4 & 64  & \checkmark \\
\texttt{MHD\_64}                            & 6 & 32  & 48 & 6  & $5\times10^{-4}$ & 1.00 & 3 & 90  & \checkmark \\
\texttt{planetswe}                          & 4 & 64  & 64 & 8  & $5\times10^{-4}$ & 0.10 & 8 & 65  & \checkmark \\
\texttt{post\_neutron\_star\_merger}        & 4 & 32  & 32 & 6  & $5\times10^{-4}$ & 0.10 & 3 & 225 & \checkmark \\
\texttt{rayleigh\_benard}                   & 6 & 64  & 96 & 6  & $1\times10^{-4}$ & 1.00 & 4 & 9   & $\times$ \\
\texttt{rayleigh\_taylor\_instability}      & 6 & 64  & 24 & 4  & $1\times10^{-4}$ & 1.00 & 6 & 48  & \checkmark \\
\texttt{shear\_flow}                        & 6 & 64  & 32 & 64 & $5\times10^{-4}$ & 0.10 & 8 & 24  & \checkmark \\
\texttt{supernova\_explosion\_64}           & 6 & 32  & 48 & 6  & $1\times10^{-4}$ & 0.10 & 3 & 24  & $\times$ \\
\texttt{turbulence\_gravity\_cooling}       & 4 & 32  & 40 & 10 & $1\times10^{-3}$ & 0.50 & 3 & 13  & $\times$ \\
\texttt{turbulent\_radiative\_layer\_2D}    & 6 & 64  & 64 & 16 & $5\times10^{-4}$ & 0.10 & 4 & 596 & \checkmark \\
\texttt{turbulent\_radiative\_layer\_3D}    & 4 & 64  & 24 & 3  & $1\times10^{-3}$ & 0.10 & 6 & 30  & $\times$ \\
\texttt{viscoelastic\_instability}          & 6 & 256 & 50 & 10 & $5\times10^{-4}$ & 1.00 & 3 & 214 & \checkmark \\
\bottomrule
\end{tabular}
}
\end{table}

\section{Dataset Descriptions and Qualitative Rollouts}
\label{app:datasets}

We evaluate 17 predictive tasks derived from the 16 base datasets in \emph{The Well}. Each subsection briefly describes the physical setting and shows a representative test trajectory, comparing ground truth with the tuned TFNO prediction at four lead times spanning the rollout. For most tasks we display one representative state variable; \texttt{rayleigh\_taylor\_instability} and \texttt{turbulence\_gravity\_cooling} include multiple fields because they are discussed separately in the main text. These figures are intended as qualitative complements to the VRMSE results rather than as an additional quantitative ranking. The code accompanying this work generates the corresponding visualizations for the remaining state variables. \\

\subsection{\texttt{acoustic\_scattering\_maze}}
A 2D acoustic wave propagation problem in which waves interact with sharply varying density structures, producing complex scattering patterns and long-range dependencies.

\wellfig{acoustic_scattering_maze}{pressure}{\gtfno{\texttt{acoustic\_scattering\_maze}}{pressure $p$}}

\clearpage

\subsection{\texttt{active\_matter}}
A 2D simulation of active particles suspended in a viscous fluid, exhibiting collective behaviour driven by long-range hydrodynamic interactions.

\hfill 

\wellfig{active_matter}{concentration}{\gtfno{\texttt{active\_matter}}{concentration $c$}}

\clearpage

\subsection{\texttt{convective\_envelope\_rsg}}
A 3D radiative hydrodynamic dataset modelling the convective envelopes of red supergiant stars, capturing large-scale stellar convection and energy transport. Three-dimensional fields are shown as a mid-depth slice.

\wellfig{convective_envelope_rsg}{velocity_r}{\gtfno{\texttt{convective\_envelope\_rsg}}{radial velocity $v_{r}$}}

\clearpage

\subsection{\texttt{euler\_multi\_quadrants}}
A 2D inviscid, compressible gas dynamics problem governed by the Euler equations with periodic boundary conditions, featuring interacting shock waves and discontinuities.

\hfill 

\wellfig{euler_multi_quadrants_periodicBC}{density}{\gtfno{\texttt{euler\_multi\_quadrants}}{density $\rho$}}

\clearpage

\subsection{\texttt{gray\_scott\_reaction\_diffusion}}
A reaction-diffusion system describing two interacting chemical species that generate self-organized Turing patterns.

\hfill 

\wellfig{gray_scott_reaction_diffusion}{A}{\gtfno{\texttt{gray\_scott\_reaction\_diffusion}}{species $A$}}

\clearpage

\subsection{\texttt{helmholtz\_staircase}}
A 2D boundary value problem involving wave propagation and scattering across stepped geometries, governed by the Helmholtz equation.

\hfill 

\wellfig{helmholtz_staircase}{pressure_re}{\gtfno{\texttt{helmholtz\_staircase}}{real part of the pressure, $\mathrm{Re}\,p$}}

\clearpage

\subsection{\texttt{MHD\_64}}
A 3D magnetohydrodynamic dataset modelling electrically conducting fluids under magnetic fields, coupling fluid motion with electromagnetic dynamics. Three-dimensional fields are shown as a mid-depth slice.

\hfill 

\wellfig{MHD_64}{magnetic_field_x}{\gtfno{\texttt{MHD\_64}}{magnetic field $B_{x}$}}

\clearpage

\subsection{\texttt{planetswe}}
A 2D shallow water simulation on a rotating spherical domain, capturing large-scale geophysical and planetary flows.

\hfill 

\wellfig{planetswe}{height}{\gtfno{\texttt{planetswe}}{height $h$}}

\clearpage

\subsection{\texttt{post\_neutron\_star\_merger}}
A 3D relativistic magnetohydrodynamic dataset describing the extreme conditions and highly nonlinear dynamics following neutron star collisions. Three-dimensional fields are shown as a mid-depth slice.

\hfill 

\wellfig{post_neutron_star_merger}{density}{\gtfno{\texttt{post\_neutron\_star\_merger}}{density $\rho$}}

\clearpage

\subsection{\texttt{rayleigh\_benard}}
A 2D natural convection problem driven by a vertical temperature gradient, producing buoyancy-induced flow structures and turbulent heat transport.

\hfill 

\wellfig{rayleigh_benard}{b}{\gtfno{\texttt{rayleigh\_benard}}{buoyancy $b$}}

\clearpage

\subsection{\texttt{rayleigh\_taylor\_instability}}
Captures the instability and mixing that arise when fluids of differing density are accelerated against one another; we evaluate the $\mathrm{At}=0.25$ configuration. Three-dimensional fields are shown as a mid-depth slice. All four published baselines fail to converge on this task (VRMSE $>10$), so all four state variables are shown here.

\hfill 

\wellfig{rayleigh_taylor_instability}{density}{\gtfno{\texttt{rayleigh\_taylor\_instability}}{density $\rho$}}
\wellfig{rayleigh_taylor_instability}{velocity_x}{\gtfno{\texttt{rayleigh\_taylor\_instability}}{velocity $v_{x}$}}
\wellfig{rayleigh_taylor_instability}{velocity_y}{\gtfno{\texttt{rayleigh\_taylor\_instability}}{velocity $v_{y}$}}
\wellfig{rayleigh_taylor_instability}{velocity_z}{\gtfno{\texttt{rayleigh\_taylor\_instability}}{velocity $v_{z}$}}

\clearpage

\subsection{\texttt{shear\_flow}}
A 2D incompressible flow exhibiting Kelvin--Helmholtz instabilities generated by velocity shear layers.

\hfill 

\wellfig{shear_flow}{tracer}{\gtfno{\texttt{shear\_flow}}{passive tracer $c$}}

\clearpage

\subsection{\texttt{supernova\_explosion\_64}}
A 3D hydrodynamic simulation of stellar collapse and shockwave propagation, modelling the explosive dynamics of supernova events. Three-dimensional fields are shown as a mid-depth slice.

\hfill 

\wellfig{supernova_explosion_64}{density}{\gtfno{\texttt{supernova\_explosion\_64}}{density $\rho$}}

\clearpage

\subsection{\texttt{turbulence\_gravity\_cooling}}
A 3D astrophysical fluid dynamics dataset incorporating self-gravity, turbulence, and radiative cooling. Three-dimensional fields are shown as a mid-depth slice. This is one of the two tasks on which the tuned TFNO does not improve on the published operator baselines (Section~\ref{sec:results}), so the three thermodynamic variables are shown here.

\hfill 

\wellfig{turbulence_gravity_cooling}{density}{\gtfno{\texttt{turbulence\_gravity\_cooling}}{density $\rho$}}
\wellfig{turbulence_gravity_cooling}{pressure}{\gtfno{\texttt{turbulence\_gravity\_cooling}}{pressure $p$}}
\wellfig{turbulence_gravity_cooling}{temperature}{\gtfno{\texttt{turbulence\_gravity\_cooling}}{temperature $T$}}

\clearpage

\subsection{\texttt{turbulent\_radiative\_layer\_2D}}
A 2D simulation of the turbulent mixing layer that forms between cold, dense gas and hot, diffuse gas, emphasizing the interaction between turbulence and radiative cooling.

\hfill 

\wellfig{turbulent_radiative_layer_2D}{density}{\gtfno{\texttt{turbulent\_radiative\_layer\_2D}}{density $\rho$}}

\clearpage

\subsection{\texttt{turbulent\_radiative\_layer\_3D}}
The 3D counterpart of the turbulent radiative layer problem, giving a more realistic representation of the mixing-layer turbulence. Three-dimensional fields are shown as a mid-depth slice.

\hfill 

\wellfig{turbulent_radiative_layer_3D}{density}{\gtfno{\texttt{turbulent\_radiative\_layer\_3D}}{density $\rho$}}

\clearpage

\subsection{\texttt{viscoelastic\_instability}}
A 2D non-Newtonian fluid dynamics dataset in which elastic stresses generate flow instabilities and chaotic behaviour even at low Reynolds numbers.

\hfill 

\wellfig{viscoelastic_instability}{c_zz}{\gtfno{\texttt{viscoelastic\_instability}}{conformation tensor $c_{zz}$}}

\end{document}